\documentclass[10pt,twocolumn,letterpaper]{article}
\usepackage{cvpr}
\usepackage{times}
\usepackage{epsfig}
\usepackage{graphicx}
\usepackage{amsmath}

\usepackage{amssymb}
\usepackage{url}            
\usepackage{amsfonts}       
\usepackage{nicefrac}       
\usepackage{microtype}      

\usepackage{bm}

\newcommand{\reals}[1]{\mathbb{R}^{#1}}
\newcommand{\hilbert}{\mathcal{H}}
\newcommand{\enorm}[1]{\left\|{#1}\right\|}
\newcommand{\hilb}[1]{\left\|{#1}\right\|_{\mathcal{H}}}
\newcommand{\myexp}[1]{e^{\left({#1}\right)}}
\newcommand{\set}[1]{\left\{#1\right\}}
\DeclareMathOperator*{\trace}{Tr}
\newcommand{\half}{\frac{1}{2}}

\renewcommand\cdots{...}

\newcommand{\suptensor}[1]{\mathfrak{S}^{d}}

\DeclareMathOperator*{\argmin}{arg\,min}
\newcommand{\fnorm}[1]{\left\|{#1}\right\|_F}

\newcommand{\comment}[1]{}

\newcommand{\ortho}{\mathcal{O}}
\newcommand{\grass}[1]{\mathcal{G}{#1}}

\DeclareMathOperator*{\subjectto}{\text{subject to}}
\DeclareMathOperator*{\sym}{sym}

\newcommand{\eye}[1]{\mathbf{I}_{#1}}
\newcommand{\mK}{\mathbf{K}}

\newcommand{\vx}{\mathbf{x}}
\newcommand{\mZ}{Z}
\newcommand{\vz}{\mathbf{z}}
\newcommand{\mX}{X}
\newcommand{\mU}{U}

\DeclareMathOperator*{\viol}{viol}
\newcommand{\kernel}{\mathbf{k}}
\newcommand{\mV}{V}
\newcommand{\mA}{A}
\newcommand{\mS}{S}
\newcommand{\mP}{P}
\newcommand{\mR}{R}
\newcommand{\mL}{L}
\newcommand{\mF}{F}
\newcommand{\ve}{\mathbf{e}}

\newcommand{\thicktilde}[1]{\mathbb{\widetilde{\text{$#1$}}}}
\newcommand{\thickhat}[1]{\mathbb{\widehat{\text{$#1$}}}}
\usepackage{subfigure}

\usepackage[pagebackref=true,breaklinks=true,letterpaper=true,colorlinks,bookmarks=false]{hyperref}

\cvprfinalcopy 

\def\cvprPaperID{3646} 

\ifcvprfinal\fi
\begin{document}

\title{Non-Linear Temporal Subspace Representations for Activity Recognition}

\author{Anoop Cherian${^{1,3}}$\quad Suvrit Sra$^2$\quad Stephen Gould$^3$\quad Richard Hartley$^3$\\
$^1$MERL, Cambridge MA,\quad $^2$MIT, Cambridge MA,\quad $^3$ACRV, ANU Canberra\\
\tt\small cherian@merl.com\quad suvrit@mit.edu\quad $\{$stephen.gould, richard.hartley$\}$@anu.edu.au
}

\maketitle
\begin{abstract}
Representations that can compactly and effectively capture the temporal evolution of semantic content are important to computer vision and machine learning algorithms that operate on multi-variate time-series data. We investigate such representations motivated by the task of human action recognition. Here each data instance is encoded by a multivariate feature (such as via a deep CNN) where action dynamics are characterized by their variations in time. As these features are often non-linear, we propose a novel pooling method, \emph{kernelized rank pooling}, that represents a given sequence compactly as the pre-image of the parameters of a hyperplane in a reproducing kernel Hilbert space, projections of data onto which captures their temporal order. We develop this idea further and show that such a pooling scheme can be cast as an order-constrained kernelized PCA objective. We then propose to use the parameters of a kernelized low-rank feature subspace as the representation of the sequences. We cast our formulation as an optimization problem on generalized Grassmann manifolds and then solve it efficiently using Riemannian optimization techniques. We present experiments on several action recognition datasets using diverse feature modalities and demonstrate state-of-the-art results. 
\end{abstract}


\section{Introduction}
\label{sec:intro}

\begin{figure}[htbp]
\includegraphics[width=9cm,trim=0cm 1cm 4cm 0cm,clip]{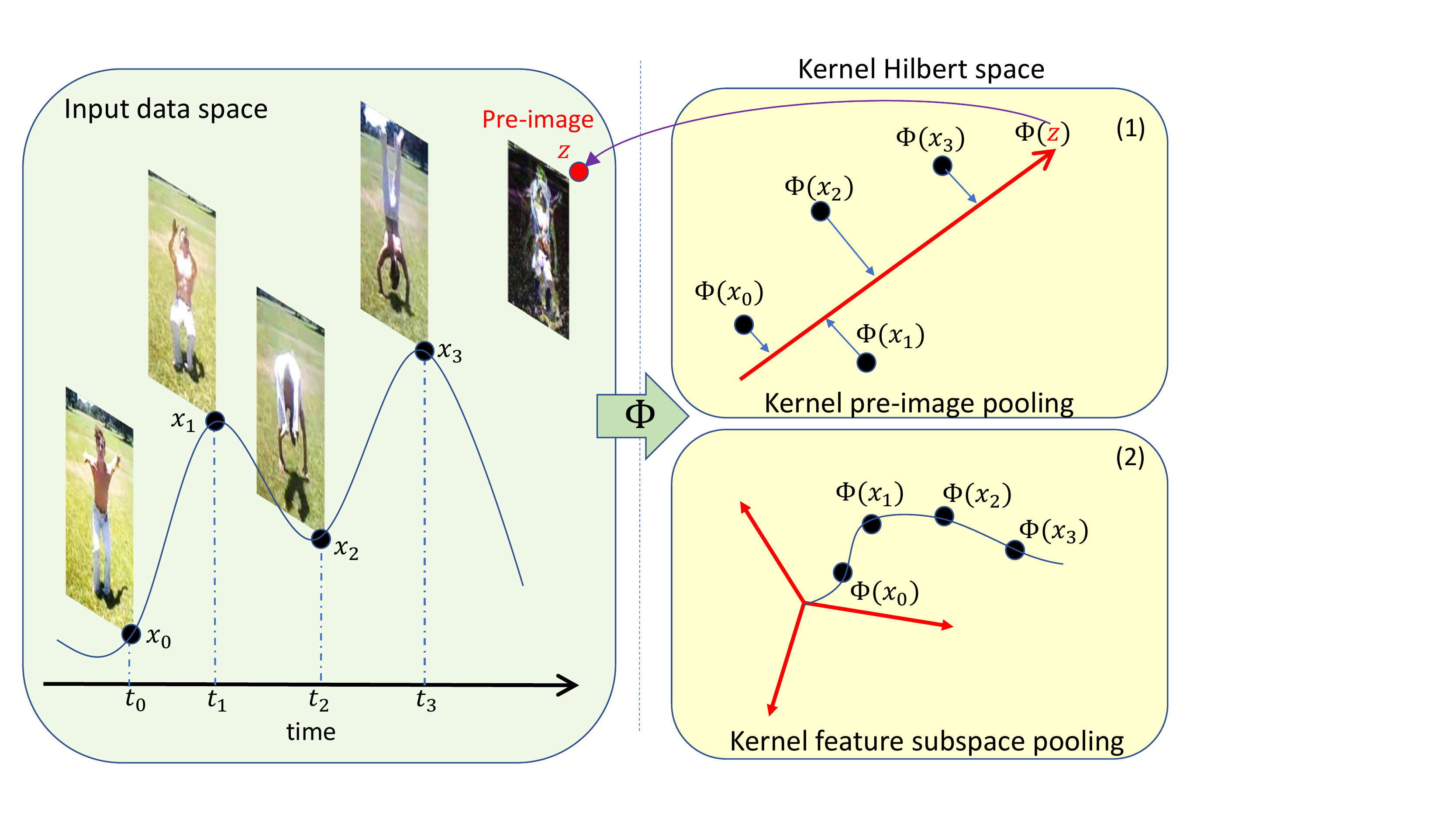}
\caption{An illustration of our two kernelized rank pooling schemes, namely (1) Pre-image pooling, that uses the pre-image $z$ (in the input space) as the pooled descriptor; this pre-image is computed such that the projections of kernel embeddings $\Phi(x)$ of input points $x$ preserve the temporal frame order when projected onto $\Phi(z)$, and (2) kernel subspace pooling, that uses the parameters of the kernel subspace for pooling, such that the projections of $\Phi(x)$ onto this subspace captures the temporal order (as increasing distances from the subspace origin). Both schemes assume that the input data is non-linear, while their kernelized embeddings (in an infinite dimensional RKHS) may allow compact linear order-preserving projections -- which can be used for pooling.}
\label{fig:krpfs_concept}
\end{figure}
In this paper, we propose compact representations for non-linear multivariate data arising in computer vision applications, by casting them in the concrete setup of action recognition in video sequences. The concrete setting we pursue is quite challenging. Although, rapid advancement of deep convolutional neural networks has led to significant breakthroughs in several computer vision tasks (e.g., object recognition~\cite{huang2017densely}, face recognition~\cite{ranjan2017hyperface}, pose estimation~\cite{wei2016convolutional}), action recognition continues to be significantly far from human-level performance. This gap is perhaps due to the spatio-temporal nature of the data and its size, which quickly outgrows processing capabilities of even the best hardware platforms. To tackle this, deep learning algorithms for video processing usually consider subsequences (a few frames) as input, extract features from such clips, and then aggregate these features into compact representations, which are then used to train a classifier for recognition.

In the popular two-stream CNN architecture for action recognition~\cite{simonyan2014two,feichtenhofer2016convolutional}, the final classifier scores are fused using a linear SVM. A similar strategy is followed by other more recent approaches such as the 3D convolutional network~\cite{tran2015learning,carreira2017quo} and temporal segment networks~\cite{wang2015action}. Given that an action is comprised of ordered variations of spatio-temporal features, any pooling scheme that discards such temporal variation may lead to sub-optimal performance. 

Consequently, various temporal pooling schemes have been devised. One recent promising scheme is \emph{rank pooling}~\cite{fernando2015modeling,Fernando:ICML2016}, in which the temporal action dynamics are summarized as the parameters of a line in the input space that preserves the frame order via linear projections. To estimate such a line, a rank-SVM~\cite{cao2007learning} based formulation is proposed, where the ranking constraints enforce the temporal order (see Section.~\ref{sec:background}). However, this formulation is limited on several fronts. First, it assumes the data belongs to a Euclidean space (and thus cannot handle non-linear geometry, or sequences of structured objects such as positive definite matrices, strings, trees, etc.). Second, only linear ranking constraints are used, however non-linear projections may prove more fruitful. Third, data is assumed to evolve smoothly (or needs to be explicitly smoothed) as otherwise the pooled descriptor may fit to random noise. 

In this paper, we introduce~\emph{kernelized rank pooling} (KRP) that aggregates data features after mapping them to a (potentially) infinite dimensional reproducing kernel Hilbert space (RKHS) via a feature map~\cite{scholkopf2001learning}. Our scheme learns hyperplanes in the feature space that encodes the temporal order of data via inner products; the pre-images of such hyperplanes in the input space are then used as action descriptors, which can then be used in a non-linear SVM for classification. This appeal to kernelization generalizes rank pooling to any form of data for which a Mercer kernel is available, and thus naturally takes care of the challenges described above. We explore variants of this basic KRP in Section~\ref{sec:krp}.

A technical difficulty with KRP is its reliance on the computation of a pre-image of a point in feature space. However, given that the pre-images are finite-dimensional representatives of infinite-dimensional Hilbert space points, they may not be unique or may not even exist~\cite{mika1998kernel}. To this end, we propose an alternative kernelized pooling scheme based on feature subspaces (KRP-FS) where instead of estimating a single hyperplane in the feature space, we estimate a low-rank kernelized subspace subject to the constraint that projections of the kernelized data points onto this subspace should preserve temporal order. We propose to use the parameters of this low-rank kernelized subspace as the action descriptor. To estimate the descriptor, we propose a novel  order-constrained low-rank kernel approximation, with orthogonality constraints on the estimated descriptor. While, our formulation looks computationally expensive at first glance, we show that it allows efficient solutions if resorting to Riemannian optimization schemes on a generalized Grassmann manifold (Section~\ref{sec:efficient_optimization}). 

We present experiments on a variety of action recognition datasets, using different data modalities, such as CNN features from single RGB frames, optical flow sequences, trajectory features, pose features, etc. Our experiments clearly show the advantages of the proposed schemes achieving state-of-the-art results. 

Before proceeding, we summarize below the main contributions of this paper.
\begin{itemize}
\setlength{\itemsep}{1pt}
\item We introduce a novel~\emph{order-constrained kernel PCA} objective that learns action representations in a kernelized feature space. We believe our formulation may be of independent interest in other applications.
\item We introduce a new pooling scheme,~\emph{kernelized rank pooling} based on kernel pre-images that captures temporal action dynamics in an infinite-dimensional RKHS. 
\item We propose efficient Riemannian optimization schemes on the generalized Grassmann manifold for solving our formulations.
\item We show experiments on several datasets demonstrating state-of-the-art results.
\end{itemize}


\section{Related Work}
\label{sec:related_work}

Recent methods for video-based action recognition use features from the intermediate layers of a CNN; such features are then pooled into compact representations. Along these lines, the popular two-stream CNN model ~\cite{simonyan2014two} for action recognition has been extended using more powerful CNN architectures incorporating intermediate feature fusion in ~\cite{feichtenhofer2016convolutional,feichtenhofer2016spatiotemporal,tran2015learning, wang_cvpr18}, however typically the features are pooled independently of their temporal order during the final sequence classification. Wang et al.~\cite{Wang2016} enforces a grammar on the two-stream model via temporal segmentation, however this grammar is designed manually. Another popular approach for action recognition has been to use recurrent networks~\cite{yue2015beyond,donahue2014long}. However, training such models is often difficult~\cite{pascanu2013difficulty} and need enormous datasets. Yet another popular approach is to use 3D convolutional filters, such as C3D~\cite{tran2015learning} and the recent I3D~\cite{carreira2017quo}; however they also demand large (and clean) datasets to achieve their best performances. 

Among recently proposed temporal pooling schemes, rank pooling~\cite{fernando2015modeling} has witnessed a lot of attention due to its simplicity and effectiveness. There have been extensions of this scheme in a discriminative setting~\cite{fernando2016discriminative,bilen2016dynamic,Fernando:ICML2016, dynamic_flow}, however all these variants use the basic rank-SVM formulation and is limited in their representational capacity as alluded to in the last section. Recently, in Cherian et al.~\cite{cherian_grp}, the basic rank pooling is extended to use the parameters of a feature subspace, however their formulation also assumes data embedded in the Euclidean space. In contrast, we generalize rank pooling to the non-linear setting and extend our formulation to an order-constrained kernel PCA objective to learn kernelized feature subspaces as data representations. To the best of our knowledge, both these ideas have not been proposed previously. 

We note that kernels have been used to describe actions earlier. For example, Cavazza et al.~\cite{cavazza2016kercov} and Quang et al.~\cite{quang2016approximate} propose kernels capturing spatio-temporal variations for action recognition, where the geometry of the SPD manifold is used for classification, and Harandi et al.~\cite{Roy_CVPR_2018,metriclearning_harandi} uses geometry in learning frameworks. Koniusz et al.~\cite{koniusz2016tensor,piotr_cvpr16,hok} uses features embedded in an RKHS, however the resultant kernel is linearized and embedded in the Euclidean space. Vemulapalli et al.~\cite{vemulapalli2014human} uses SE(3) geometry to classify pose sequences. Tseng~\cite{Tseng2012} proposes to learn a low-rank subspace where the action dynamics are linear. Subspace representations have also been investigated~\cite{turaga2011statistical,harandi2013kernel,OHara2012}, and the final representations are classified using Grassmannian kernels.  However, we differ from all these schemes in that our subspace is learned using temporal order constraints, and our final descriptor is an element of the RKHS, offering greater flexibility and representational power in capturing non-linear action dynamics. We also note that there have been extensions of kernel PCA for computing pre-images, such as for denoising~\cite{mika1998kernel,kwok2004pre}, voice recognition~\cite{kwok2004eigenvoice}, etc., but are different from ours in methodology and application.

\section{Preliminaries}
\label{sec:background}
In this section, we setup the notation for the rest of the paper and review some prior formulations for pooling multivariate time series for action recognition. Let $\mX=\left[\vx_1, \vx_2,\cdots, \vx_n\right]$ be features from $n$ consecutive frames of a video sequence, where we assume each $\vx_i\in\reals{d}$. 

Rank pooling~\cite{fernando2015modeling} is a scheme to compactly represent a sequence of frames into a single feature that summarizes the sequence dynamics. Typically, rank pooling solves the following objective:
\begin{equation}
\label{eq:rank-basic}
\min_{\vz\in\reals{d}} \half\enorm{\vz}^2 + \lambda\sum_{i<j} \max(0, \eta + \vz^T \vx_i -\vz^T \vx_j),
\end{equation}
where $\eta > 0$ is a threshold enforcing the temporal order and $\lambda$ is a regularization constant. Note that, the formulation in~\eqref{eq:rank-basic} is the standard Ranking-SVM formulation~\cite{cao2007learning} and hence the name. The minimizing  vector $\vz$ (which captures the parameters of a line in the input space) is then used as the pooled action descriptor for $\mX$ in a subsequent classifier. The rank pooling formulation in~\cite{fernando2015modeling} encodes the temporal order as increasing intercept of input features when projected on to this line. 

The objective in~\eqref{eq:rank-basic} only considers preservation of the temporal order; as a result, the minimizing $\vz$ may not be related to 
input data at a semantic level (as there are no constraints enforcing this). It may be beneficial for $\vz$ to capture some discriminative properties of the data (such as human pose, objects in the scene, etc.), that may help a subsequent classifier. To account for these shortcomings, Cherian et al.~\cite{cherian_grp}, extended rank pooling to use the parameters of a subspace as the representation for the input features with better empirical performance. Specifically,~\cite{cherian_grp} solves the following problem.
\begin{align}
\label{eq:low-rank}
\min_{\mU\in\grass(p,d)} &\half\sum_{i=1}^n\enorm{\vx_i - \mU\mU^T\vx_i}^2 + \notag\\
&\quad\sum_{i<j}\max(0, \eta + \enorm{\mU^T \vx_i}^2 - \enorm{\mU^T \vx_j}^2),
\end{align}
where instead of a single $\vz$ as in~\eqref{eq:rank-basic}, they learn a subspace $\mU$ (belonging to a $p$-dimensional Grassmann manifold embedded in $\reals{d}$), such that this $\mU$ provides a low-rank approximation to the data, as well as, projection of the data points onto this subspace will preserve their temporal order in terms of their distance from the origin.

However, both the above schemes have limitations; they assume input data is Euclidean, which may be severely limiting when working with features from an inherently non-linear space. To circumvent these issues, in this paper, we explore kernelized rank pooling schemes that generalize rank pooling to data objects that may belong to any arbitrary geometry, for which a valid Mercer kernel can be computed. As will be clear shortly, our schemes generalize both~\cite{fernando2015modeling} and~\cite{cherian_grp} as special cases when the kernel used is a linear one. In the sequel,  we assume an RBF kernel for the feature map, defined for $\vx,\vz\in\reals{d}$ as: $\kernel(\vx, \vz) = \exp\left\{-\frac{\enorm{\vx-\vz}^2}{2\sigma^2}\right\}$,
for a bandwidth $\sigma$. We use $\mK$ to denote the $n\times n$ RBF kernel matrix constructed on all frames in $\mX$, i.e., the $ij$-th element $\mK_{ij} = \kernel(\vx_i, \vx_j)$, where $\vx_i,\vx_j\in\mX$. Further, for the kernel $\kernel$, let there exist a corresponding feature map $\Phi: \reals{d}\times\reals{d}\to\hilbert$, where $\hilbert$ is a Hilbert space for which 
$\langle \Phi(\vx), \Phi(\vz)\rangle_{\hilbert} = \kernel(\vx, \vz)$,
for all $\vx,\vz\in\reals{d}$.
\section{Our Approach}
\label{sec:krp}
Given a sequence of temporally-ordered features~$\mX$, our main idea is to use the kernel trick to map the features to a (plausibly) infinite-dimensional RKHS~\cite{scholkopf2001learning}, in which the data is linear. We propose to learn a hyperplane in the RKHS, projections of the data to which will preserve the temporal order. We formalize this idea below and explore variants.

\subsection{Kernelized Rank Pooling}
\label{sec:bkrp}
For data points $\vx\in X$ and their RKHS embeddings $\Phi(\vx)$, a straightforward way to extend~\eqref{eq:rank-basic} is to use a direction $\Phi(\vz)$ in the feature space, projections of $\Phi(\vx)$ onto this line will preserve the temporal order. However, given that we need to retrieve $\vz$ in the input space, to be used as the pooled descriptor, one way (which we propose) is to compute the pre-image $\vz$ of $\Phi(\vz)$, which can then used as the action descriptor in a subsequent non-linear action classifier. Mathematically, this~\emph{basic kernelized rank pooling} (BKRP) formulation is:
\begin{align}
&\arg\min_{\vz\in\reals{d}}\quad \text{BKRP}(\vz):= \half\enorm{\vz}^2 + \notag\\
&\quad \lambda\sum_{i<j} \max(0, \eta + \langle \Phi(\vx_i), \Phi(\vz)\rangle - \langle \Phi(\vx_j), \Phi(\vz)\rangle \\
&= \half\enorm{\vz}^2 + \lambda \sum_{i<j} \max(0, \eta + \kernel(\vx_i, \vz) - \kernel(\vx_j, \vz)).
\label{eq:bkrp}
\end{align}

As alluded to earlier, a technical issue with~\eqref{eq:bkrp} (and also with~\eqref{eq:rank-basic}) is that the optimal direction $\vz$ may ignore any useful properties of the original data $\mX$ (instead could be some line that preserves the temporal order alone). To make sure $\vz$ is similar to $\vx\in\mX$, we write an improved~\eqref{eq:bkrp} as:
\begin{align}
&\argmin_{\vz\in\reals{d},\xi\geq 0} \text{IBKRP}(\vz) := \half\sum_{i=1}^n\enorm{\vx_i-\vz}^2 + C\sum_{i,j=1}^n\xi_{ij}\notag\\
&+\lambda\sum_{i<j} \max(0, \eta-\xi_{ij} + \kernel\left(\vx_i, \vz) - \kernel(\vx_j, \vz)\right),
\end{align}
where the first component says that the computed pre-image is not far from the input data.\footnote{When $\vx$ is not an object in the Euclidean space, we assume $\enorm{ . }$ to define some suitable distance on the data.} The variables $\xi$ represent non-negative slacks and $C$ is a positive constant. 

The above formulation assumes a pre-image $\vz$ always exists, which may not be the case in general, or may not be unique even if it exists~\cite{mika1998kernel}. We could avoid this problem by simply not computing the pre-image, instead keeping the representation in the RKHS itself. To this end, in the next section, we propose an alternative formulation in which we assume useful data maps to a $p$-dimensional subspace of the RKHS, in which the temporal order is preserved. We propose to use the parameters of this subspace as our sequence representation. Compared to a single $\vz$ to capture action dynamics (as described above), a subspace offers a richer representation (as is also considered in~\cite{cherian_grp})

\subsection{Kernelized Subspace Pooling}
Reusing our notation, let $\vx_1, \vx_2,\cdots, \vx_n$ be the $n$ points in $\reals{d}$ from a sequence. Since it is difficult to work in the complete Hilbert space $\hilbert$, we restrict ourselves to the subspace of $\hilbert$ spanned by the $\Phi(\vx_i), i=1,2,\cdots, n$. For convenience, let this space be called $\hilbert$. Assuming that they are all linearly independent, the Representer theorem~\cite{wahba1990spline,scholkopf2001generalized} says that the points $\set{\Phi(\vx_i)}$ can be chosen as a basis for $\hilbert$ (not in general an orthonormal basis, however). In this case, $\hilbert$ is a space of dimension $n$. 

As alluded to above, we are concerned with the case where all the $\Phi(\vx_i)$ may lie close to some $p$-dimensional subspace of $\hilbert$, denoted by $V$. This space will initially be unknown, and is to be determined in some manner. Denote by $\Omega_p:\hilbert\to V$, the orthogonal projection from $\hilbert$ to $V$. Assume that $V$ has an orthonormal basis, $\set{\ve_i\ |\ i=1,\cdots,p}$. In terms of the basis $\set{\Phi(\vx_j)}$ for $\hilbert$, we can write 
\begin{equation}
\ve_i = \sum_{j=1}^n a_{ij}\Phi(\vx_j),
\label{eq:1}
\end{equation}
for appropriate scalars $a$.  Let $\Omega_p(\Phi(\vx))$ denote the embedding of the input data point $\vx\in\reals{d}$ in this kernelized subspace. Then,
\begin{equation}
\Omega_p(\Phi(\vx)) = \sum_{i=1}^p \langle \Phi(\vx), \ve_i\rangle\ \ve_i.
\label{eq:2}
\end{equation}
Substituting~\eqref{eq:1} into~\eqref{eq:2}, we obtain
\begin{align}
\Omega_p(\Phi(\vx))\!&=\!\!\sum_{i=1}^p\sum_{j=1}^n a_{ij}\langle\Phi(\vx), \Phi(\vx_j)\rangle \left(\sum_{\ell=1}^n a_{i\ell}\Phi(\vx_{\ell})\right)\notag\\
&= \sum_{i=1}^p\sum_{j=1}^n\sum_{\ell=1}^nk(\vx, \vx_j) a_{ji}^T a_{i\ell} \Phi(\vx_{\ell}),
\end{align}

\noindent which can be written using matrix notation as:
\begin{equation}
\Omega_p(\Phi(\vx)) = \Phi(\mX)\mA\mA^T\kernel(\mX, \vx),
\label{eq:omega}
\end{equation}
where $\kernel(\mX, \vx)$ denotes the $n\times 1$ vector whose $j$-th dimension is given by $\kernel(\vx_j,\vx)$ and $\mA$ is an $n\times p$ matrix. Note that there is a certain abuse of notation here in that $\Phi(\mX)$ is a matrix with $n$ columns, each column of which is an element in $\hilbert$, where as the other matrices are those of real numbers. 


Using these notation, we propose a novel kernelized feature subspace learning (KRP-FS) formulation below, with ordering constraints: 
\begin{align}
\label{eq:krpfs-obj}
&\arg\min_{\mA\in\reals{n\times p}}\!\!\!\mF(\mA) := \half\sum_{i=1}^n   \hilb{\Phi(\vx_i) - \Omega_p(\Phi(\vx_i))}^2 \\
&\text{s.t.}\ \hilb{\Omega_p(\Phi(\vx_i))}^2 + \eta \leq \hilb{\Omega_p(\Phi(\vx_j))}^2,\ \forall i\!<\!j,
\label{eq:krpfs-cons}
\end{align}
where~\eqref{eq:krpfs-obj} learns a $p$-dimensional feature subspace of the RKHS in which most of the data variability is contained, while~\eqref{eq:krpfs-cons} enforces the temporal order of data in this subspace, as measured by the length (using a Hilbertian norm $\hilb{.}$) of the projection of the data points $\vx$ onto this subspace. Our main idea is to use the parameters of this kernelized feature subspace projection $\Omega_p$ as the pooled descriptor for our input sequence, for subsequent sequence classification. To ensure that such descriptors from different sequences are comparable\footnote{Recall that such normalization is common when using vectorial representations, in which case they are unit normalized.}, we need to ensure that $\Omega_p$ is normalized, i.e., has orthogonal columns in the feature space, viz., $\Omega_p^T \Omega_p= \eye{p}$. In the view of~\eqref{eq:1}, this implies the basis $\set{\ve_i}$ for $V$ satisfies $\langle \ve_i, \ve_j \rangle =\delta_{ij} $ (the delta function) and boils down to:
\begin{equation}
\Omega_p^T\Omega_p = \mA^T\mK\mA = \eye{p},
\label{eq:gengrass}
\end{equation}
where $\mK$ is the kernel constructed on $\mX$ and is symmetric positive definite (SPD). Incorporating these conditions and including slack variables (to accommodate any outliers) using a regularization constant $C$, we rewrite~\eqref{eq:krpfs-obj},~\eqref{eq:krpfs-cons} as:
\begin{align}
\label{eq:krpfs-obj-1}
&\argmin_{\small{\substack{\mA\in\reals{n\times p},\\\mA^T\mK\mA=\eye{p}, \xi\geq 0}}}\hspace*{-0.3cm}\mF(\mA)\!\!:=\!\!\half\!\!\sum_{i=1}^n\!\hilb{\Phi(\vx_i)\!-\! \Omega_p(\Phi(\vx_i))}^2\!\!+\!C\!\sum_{i<j}^n\!\!\xi_{ij} &\\
&\text{s.t.} \hilb{\Omega_p(\Phi(\vx_i))}^2\!\!+\!\!\eta\!-\xi_{ij}\!\leq\!\!\hilb{\Omega_p(\Phi(\vx_j))}^2,\!\!\forall i<j.
\label{eq:krpfs-cons-1}
\end{align}

It may be noted that our objective~\eqref{eq:krpfs-obj-1} essentially depicts kernel principal components analysis (KPCA)~\cite{scholkopf1997kernel}, albeit the constraints make estimation of the low-rank subspace different, demanding sophisticated optimization techniques for efficient solution. We address this concern below, by making some key observations regarding our objective.

\subsection{Efficient Optimization}
\label{sec:efficient_optimization}
Substituting for the definitions of $\mK$ and $\Omega_p$, the formulation in~\eqref{eq:krpfs-obj-1} can be written using hinge-loss as:
\begin{align}
&\argmin_{\substack{\mA\in\reals{n\times p}| \mA^T\mK\mA = \eye{p}\\ \xi\geq 0}}\hspace*{-0.7cm}\mF(\mA) := \half\sum_{i=1}^n -2\kernel(\mX,\vx_i)^T\mA\mA^T\kernel(\mX,\vx_i) \notag\\
&\qquad+\kernel(\mX,\vx_i)^T\mA\mA^T \mK \mA\mA^T\kernel(\mX, \vx_i)+C\sum_{i<j}\xi_{ij}\nonumber\\ 
&\qquad + \lambda\sum_{i<j} \max\bigg(0, \kernel(\mX, \vx_i)^T\mA\mA^T\mK\mA\mA^T\kernel(\mX,\vx_i)+ \notag\\
&\qquad-\kernel(\mX,\vx_j)^T\mA\mA^T\mK\mA\mA^T\kernel(\mX,\vx_j)\!\!+\!\!\eta-\!\!\xi_{ij}\bigg).
\label{eq:krpfs-hinge-loss}
\end{align}
As is clear, the variable $\mA$ appears as $\mA\mA^T$ through out and thus our objective is invariant to any right rotations by an element of the $p$-dimensional orthogonal group $\ortho(p)$, i.e., for $\mR\in\ortho(p), \mF(\mA) = \mF(\mA\mR)$. This, together with the condition in~\eqref{eq:gengrass} suggests that the optimization on $\mA$ as defined in~\eqref{eq:krpfs-hinge-loss} can be solved over the so called~\emph{generalized Grassmann manifold}~\cite{edelman1998geometry}[Section 4.5] using Riemannian optimization techniques. 

We use a Riemannian conjugate gradient (RCG) algorithm~\cite{smith1994optimization} on this manifold for solving our objective. A key component for this algorithm to proceed is the expression for the Riemannian gradient of the objective $\mathrm{grad}_{A} \mF(\mA)$, which for a generalized Grassmannian can be obtained from the Euclidean gradient $\nabla_{\mA}\mF(\mA)$ as:
\begin{equation}
\mathrm{grad}_\mA\mF(\mA) = \mK^{-1}\nabla_{\mA}\mF(\mA)- \mA\sym(A^T\nabla_{\mA} \mF(\mA)),
\end{equation}
where $\sym(\mL) = \frac{L+L^T}{2}$ is the symmetrization operator~\cite{mishra2016riemannian}[Section 4]. The Euclidean gradient for~\eqref{eq:krpfs-hinge-loss} is as follows: let $\mS_1=\mK\mK\mA$, $\mS_2=\mK\mA\mA^T$, and $\mS_3=\mA^T\mK\mA$, then
\begin{equation}
\nabla_{\mA}\mF(\mA) = \mS_1\left(\mS_3\!-\!2\eye{p}\right)\!+\!\mS_2\mS_1\!+\!\!\lambda\left(\mK_{12}\mA\mS_3\!+\!\mS_2\mK_{12}\mA\right),
\label{eq:euc_grad}
\end{equation}
where $\mK_{12}= \mK_1\mK_1^T - \mK_2\mK_2^T$, $\mK_1,\mK_2$ are the kernels capturing the order violations in~\eqref{eq:krpfs-hinge-loss} for which the hinge-loss is non-zero; $\mK_1$ collecting the sum of all violations for $\vx_i$ and $\mK_2$ the same for $\vx_j$. If further scalability is desired, one can also invoke stochastic Riemannian solvers such as Riemannian-SVRG~\cite{zhang2016riemannian,kasai2016riemannian} instead of RCG. These methods extend the variance reduced stochastic gradient methods to Riemannian manifolds, and may help scale the optimization to larger problems.

\subsection{Action Classification}
Once we find $\mA$ per video sequence by solving~\eqref{eq:krpfs-hinge-loss}, we use $\Omega = \Phi(\mX)\mA$ (note that we omit the subscript $p$ from $\Omega_p$ as it is by now understood) as the action descriptor. However, as $\Omega$ is semi-infinite, it cannot be directly computed, and thus we need to resort to the kernel trick again for measuring the similarity between two encoded sequences. Given that $\Omega\in\grass(p)$ belong to a generalized Grassmann manifold, we can use any Grassmannian kernel for computing this similarity. Among several such kernels reviewed in~\cite{harandi2014expanding}, we found the exponential projection metric kernel to be empirically beneficial. For two sequences $\mX_1\in\reals{d\times n_1},\mX_2\in\reals{d\times n_2}$, their subspace parameters $\mA_1,\mA_2$ and their respective KRP-FS descriptors $\Omega_1,\Omega_2\in\grass(p)$, the exponential projection metric kernel is defined as:
\begin{equation}
\mathbb{K}_{\mathcal{G}}(\Omega_1, \Omega_2) = \exp\left(\nu \fnorm{\Omega_1^T\Omega_2}^2\right), \text{for } \nu>0.
\end{equation}
Substituting for $\Omega$'s, we have the following kernel for action classification, whose $ij$-th entry is given by:
\begin{align}
\mathbb{K}_{\mathcal{G}}^{ij}(\Omega_1,\Omega_2) &= \exp\left(\nu\fnorm{\mA_i \mK \mA_j}^2\right),
\end{align}
where $\mK\in\reals{n_1\times n_2}$ is an (RBF) kernel capturing the similarity between sequences. 

\subsection{Nystr\"om Kernel Approximation}
A challenge when using our scheme on large datasets is the need to compute the kernel matrix (such as in a non-linear SVM); this computation can be expensive. To this end, we resort to the well-known Nystr\"om-based low-rank kernel approximations~\cite{drineas2005nystrom}. Technically, in this approximation, only a few columns of the kernel matrix are computed, and the full kernel is approximated by a low-rank outer product. In our context, let $\thicktilde{\mathbb{K}}_{\mathcal{G}}\in\reals{n\times m}$ (for $m<<n$) represents a matrix with $m$ randomly chosen columns of $\mathbb{K}_{\mathcal{G}}$, then the Nystr\"om approximation $\thickhat{\mathbb{K}}_{\mathcal{G}}$ of $\mathbb{K}_{\mathcal{G}}$ is given by:
\begin{equation}
\thickhat{\mathbb{K}}_{\mathcal{G}} = \thicktilde{\mathbb{K}}_{\mathcal{G}}M\thicktilde{\mathbb{K}}_{\mathcal{G}}^T,
\end{equation}
where $M$ is the (pseudo) inverse of the first $m\times m$ sub-matrix of $\thicktilde{K}_{\mathcal{G}}$. Typically, only a small fraction (1/8-th of $\mathbb{K}_{\mathcal{G}}$ in our experiments, Table~\ref{tab:hmdb_nystrom}) of columns are needed to approximate the kernel without any significant loss in performance.

\section{Computational Complexity}
Evaluating the objective in~\eqref{eq:krpfs-hinge-loss} takes $\mathcal{O}(n^3p + m^3 + mnd)$ operations and computing the Euclidean gradient in~\eqref{eq:euc_grad} needs $\mathcal{O}(n^3+n^2p)$ computations for each iteration of the solver. While, this may look more expensive than the basic ranking pooling formulation, note that here we use kernel matrices, which for action recognition datasets, are much smaller in comparison to very high dimensional (CNN) features used for frame encoding. See Table~\ref{tab:runtime} for empirical run-time analysis. 

\section{Experiments}
\label{sec:expts}
In this section, we provide experiments on several action recognition datasets where action features are represented in diverse ways. Towards this end, we use (i) the JHMDB and MPII Cooking activities datasets, where frames are encoded using a VGG two-stream CNN model, (ii) HMDB dataset, for which we use features from a ResNet-152 CNN model, (iii) UTKinect actions dataset, using non-linear features from 3D skeleton sequences corresponding to human actions, and (iv) hand-crafted bag-of-words features from the MPII dataset. For all these datasets, we compare our methods to prior pooling schemes such as average pooling (in which features per sequence are first averaged and then classified using a linear SVM), rank pooling (RP)~\cite{fernando2015modeling}, and generalized rank pooling (GRP)~\cite{cherian_grp}. We use the publicly available implementations of these algorithms without any modifications. Our implementations are in Matlab and we use the ManOpt package~\cite{boumal2014manopt} for Riemannian optimization. Below, we detail our datasets and their pre-processing, following which we furnish our results. 

\subsection{Datasets and Feature Representations}
\paragraph*{{HMDB Dataset~\cite{kuehne2011hmdb}}:}is a standard benchmark for action recognition, consisting of 6766 videos and 51 classes. The standard evaluation protocol is three-split cross-validation using classification accuracy. To generate features, we use a standard two-stream ResNet-152 network pre-trained on the UCF101 dataset (available as part of~\cite{feichtenhofer2016spatiotemporal}). We use the 2048D pool5 layer output to represent per-frame features for both streams. 

\noindent\textbf{{JHMDB Dataset~\cite{jhuang2013towards}}:} consists of 960 video sequences and 21 actions. The standard evaluation protocol is average classification accuracy over three-fold cross-validation. For feature extraction, we use a two-stream model using a VGG-16 network. To this end, we fine-tune a network, pre-trained on the UCF101 dataset (provided as part of~\cite{feichtenhofer2016convolutional}).  We extract 4096D fc6 layer features as our feature representation. 

%
\noindent\textbf{{MPII Cooking Activities Dataset~\cite{rohrbach2012database}}:} consists of cooking actions of 14 individuals. The dataset has 5609 video sequences and 64 different actions. For feature extraction, we use the fc6 layer outputs from a two-stream VGG-16 model. We also present experiments with dense trajectories (encoded as bag-of-words using 4K words). We report the mean average precision over 7 splits.

\noindent\textbf{{UTKinect Actions~\cite{xia2012view}}:} is a dataset for action recognition from 3D skeleton sequences; each sequence has 74 frames. There are 10 actions in the dataset performed by 2 subjects. We use this dataset to demonstrate the effectiveness of our schemes to explicit non-linear features. We encode each 3D pose using a Lie algebra based scheme that maps the skeletons into rotation and translation vectors which are objects in the SE(3) geometry as described in~\cite{vemulapalli2014human}. We report the average classification accuracy over 2 splits.
\\

\begin{figure*}[t]
\center
\subfigure[]{\label{fig:thresh_flow}\includegraphics[width=4cm,trim=0cm 7.2cm 0.5cm 7.8cm, clip]{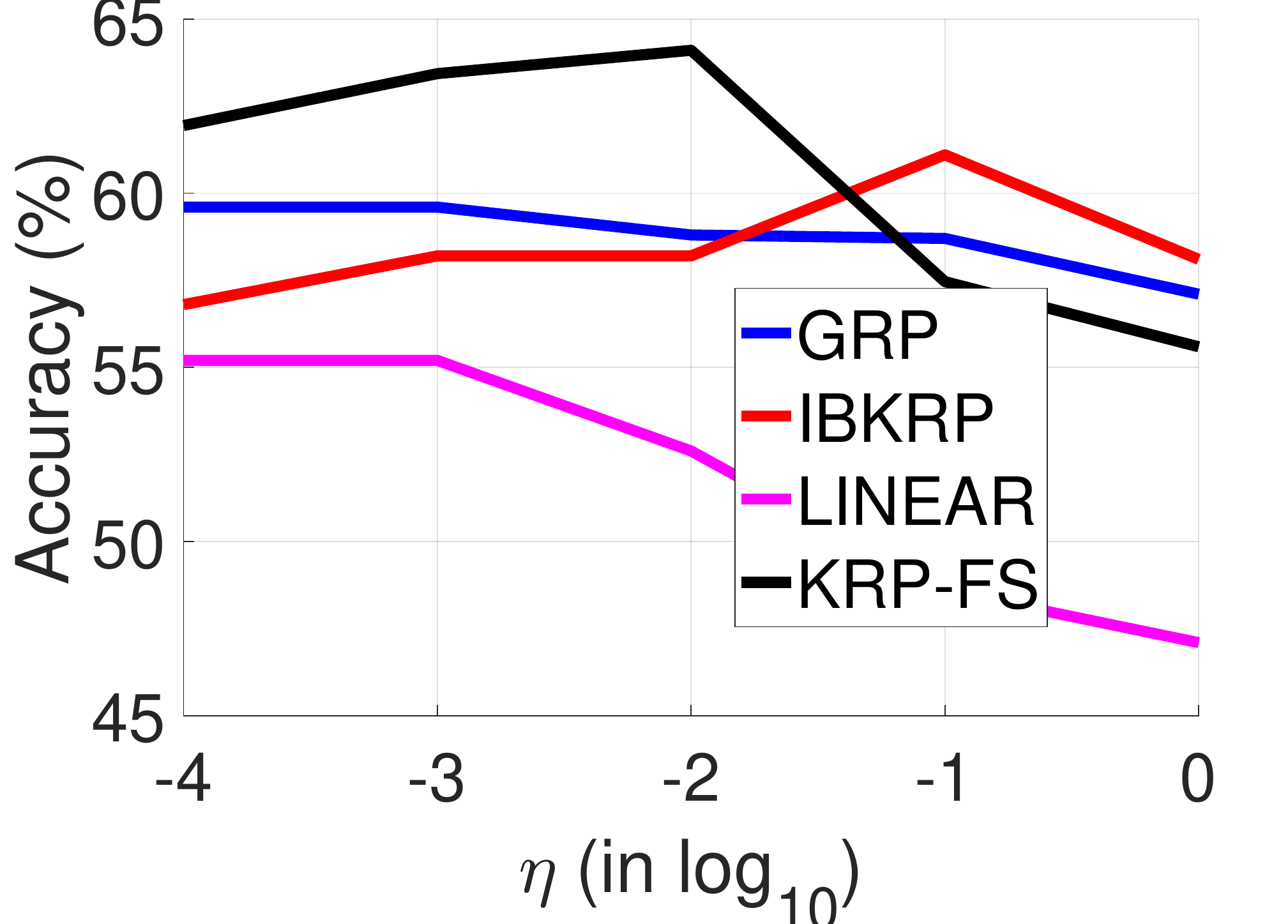}}
\subfigure[]{\label{fig:thresh_rgb}\includegraphics[width=4cm,trim=0cm 7.2cm 0.5cm 7.8cm, clip]{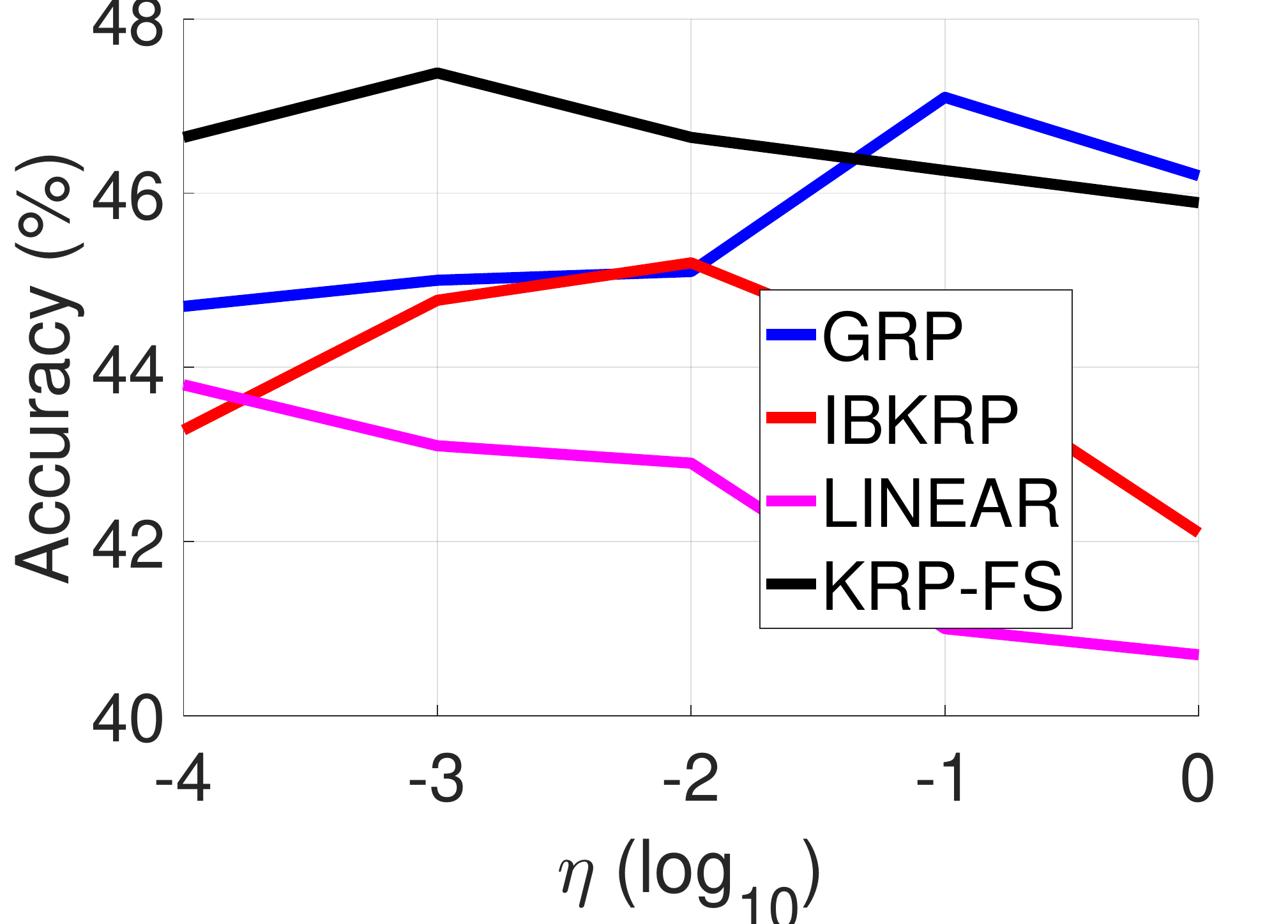}}
\subfigure[]{\label{fig:subspaces}\includegraphics[width=4cm,trim=0cm 7.1cm 0.5cm 7.8cm, clip]{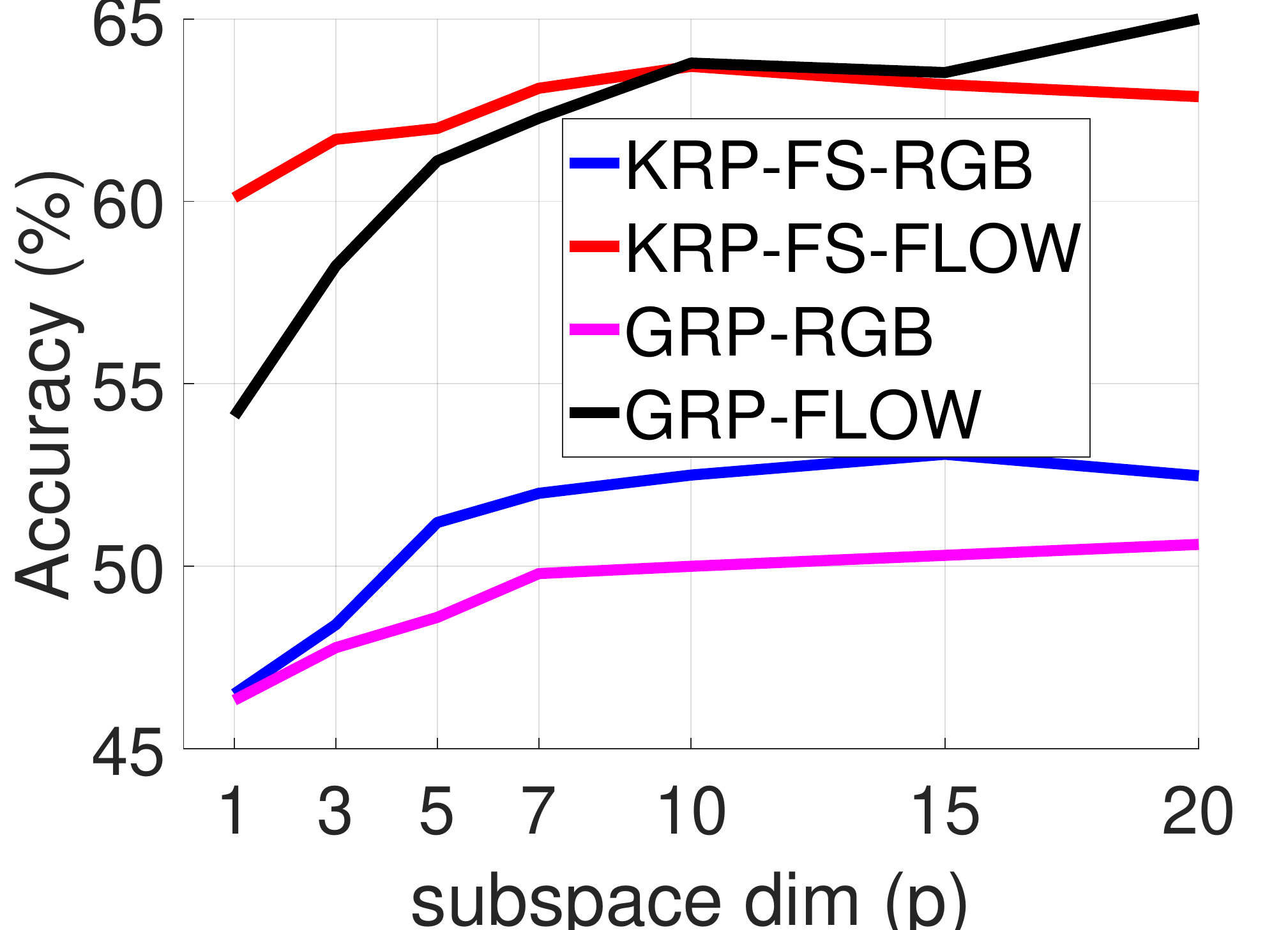}}
\subfigure[]{\label{fig:pca_dim}\includegraphics[width=4cm,trim=0cm 7.2cm 0.5cm 7.8cm, clip]{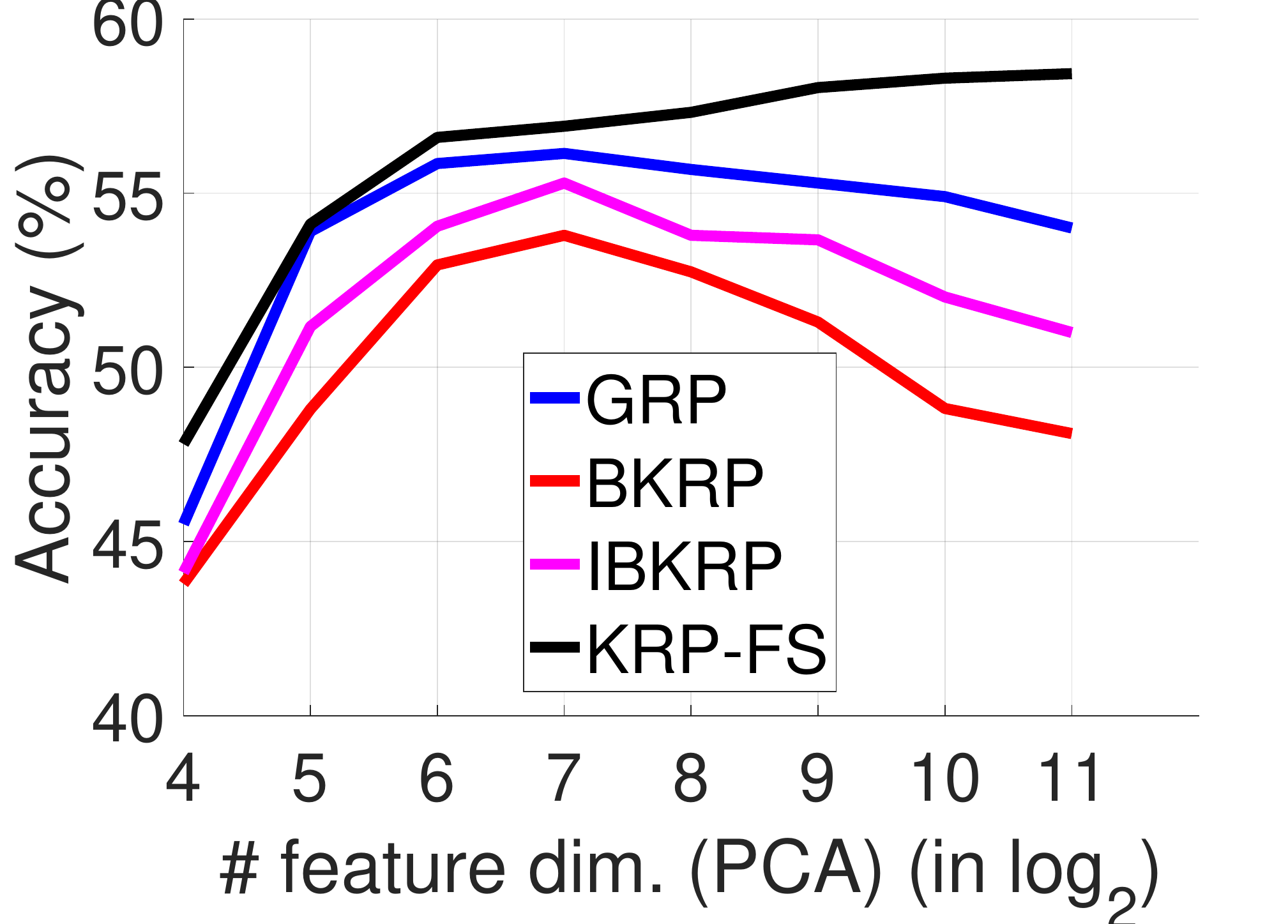}}
\caption{Analysis of parameters in our schemes: (a,b) accuracy against increasing ordering threshold $\eta$ for RGB and FLOW streams respectively on JHMDB dataset split-1, (c) classification accuracy against increasing subspace dimensionality on HMDB-51 split1, and (d) Effect of applying PCA to input features before using our schemes (on HMDB split1).}
\end{figure*}

\subsection{Analysis of Model Parameters}
\noindent\textbf{Ranking Threshold $\eta$:} An important property of our schemes to be verified is whether it reliably detects temporal order in the extracted features and if so how does the ordering parameter $\eta$ influence the generated descriptor for action classification. In Figures~\ref{fig:thresh_flow},~\ref{fig:thresh_rgb}, we plot the accuracy against increasing values of the ranking threshold $\eta$ for features from the RGB and flow streams on JHMDB split-1. The threshold $\eta$ was increased from $10^{-4}$ to 1 at multiples of 10. We see from the Figure~\ref{fig:thresh_flow} that increasing $\eta$ does influence the nature of the respective algorithms, however each algorithm appears to have its own setting that gives the best performance;e.g., IBKRP achieves best at $\eta=0.1$, while KRP-FS takes the best value at $\eta=0.0001$. We also plot the same for GRP~\cite{cherian_grp} and a linear kernel; the latter takes a dip in accuracy as $\eta$ increases, because higher values of $\eta$ are difficult to satisfy for our unit norm features when only linear hyperplanes are used for representation. In Figure~\ref{fig:thresh_rgb}, we see a similar trend for the RGB stream. 
\\
\noindent\textbf{Number of Subspaces $p$:} In Figure~\ref{fig:subspaces}, we plot the classification accuracy against increasing number of subspaces on the HMDB dataset split-1. We show the results for both optical flow and image streams separately and also plot the same for GRP (which is a linear variant of KRP-FS). The plot shows that using more subspaces is useful, however beyond say 10-15, the performance saturates suggesting perhaps temporally ordered data inhabits a low-rank subspace, as was the original motivation to propose this idea.

\noindent\textbf{Dimensionality Reduction:} To explore the possibility of dimensionality reduction (using PCA) of the CNN features and understand how well the kernelization results be (does the lower-dimensional features still capture the temporal cues?), in Figure~\ref{fig:pca_dim}, we plot accuracy against increasing dimensionality in PCA on 2048D ResNet-152 features from HMDB-51 split1. As the plot shows, for almost all the variants of our methods, using a reduced dimension does result in useful representations -- perhaps because we remove noisy dimensions in this process. We also witness that KRP-FS performs the best in generating representations from very low-dimensional features. 
\\
\noindent\textbf{Nystr\"om Approximation Quality:} In Table~\ref{tab:hmdb_nystrom}, we plot the kernel approximation quality using Nystr\"om on HMDB split-1 features for the KRP-FS variant. We uniformly sampled data in the order of $1/2^k$-th original data size, where $k$ is varied from 5 to 2. We see from the table that the accuracy decreases only marginally ($< 1$\%) with the approximation. In the sequel, we use a factor of 1/8-th data size. 

\noindent\textbf{Runtime Analysis:} For this experiment, we use a single core Intel 2.6 GHz CPU. To be compatible with others, we re-implemented~\cite{fernando2015modeling} as a linear kernel. Table~\ref{tab:runtime} presents the results. As expected, our methods are slightly costlier than others, mainly due to the need to compute the kernel. However, they are still fast, and even with our unoptimized Matlab implementation, could run at real-time rates (27fps). Further, increasing the number of subspaces in KRP-FS does not make a significant impact on the running time; e.g., (increasing from 3 to 20 increased by 1.2ms).

\noindent\textbf{Feature Pre-processing:} As described in~\cite{fernando2015modeling} and~\cite{fernando2016discriminative}, taking a signed-square root (SSR) and temporal moving-average (MA) of the features improve accuracy. In Table~\ref{tab:ma-perf}, we revisit these steps in the kernelized setting (KRP-FS) using 3 subspaces. It is clear these steps bring only very marginal improvements. This is unsurprising; as is known RBF kernel already acts as a low-pass filter.

\noindent\textbf{Homogeneous Kernels:}
A variant of rank pooling~\cite{fernando2015modeling} uses homogeneous kernels~\cite{vedaldi2012efficient} to map data into a linearizable Hilbert space onto which linear rank pooling is applied. In Table~\ref{tab:chisq-hmdb-results}, we use a Chi-squared kernel for rank pooling and compare the performance to BKRP (using RBF kernel). While we observe a 5\% improvement over~\cite{fernando2015modeling}  when using homogeneous kernels, still BKRP (which is more general) significantly outperforms it.

\begin{table}[htbp]
	\centering
	\begin{tabular}{c|c|c|c}
        \hline
		HMDB Dataset &  FLOW  & RGB & FLOW+RGB \\				
		\hline
        RP~\cite{fernando2015modeling} & 56.7 &  38.3 & 63.1  \\
        Hom. Chi-Sq. RP~\cite{fernando2015modeling} & 61.5 & 43.8 & 66.5 \\
        \hline
        BKRP   (ours)               & 54.9 &  45.9 & 69.5 \\
	\end{tabular}
	\caption{Comparisons to rank pooling using a homogeneous kernel linearization of CNN features via a Chi-Squared kernel as in~\cite{fernando2015modeling}.}          
    \label{tab:chisq-hmdb-results}
\end{table}

\begin{table}[htbp]
\centering
  \begin{tabular}{c|c|c}
    \hline
    Method & FLOW & RGB \\
    \hline
    with MA + SSR & 61.5 & 51.6\\
    w/o MA + SSR & 61.4 & 51.4\\
    w/o MA + w/o SSR & 60.8 & 51.3 \\
  \end{tabular}
\caption{Effect of Moving Average (MA) and signed-square root (SSR) of CNN features before KRP-FS on HMDB split-1.}
\label{tab:ma-perf}
\end{table}

\subsection{Comparisons between Pooling Schemes}
In Tables~\ref{tab:jhmdb-split1-results},~\ref{tab:hmdb-split1-results}, and~\ref{tab:mpii-split1-results}, we compare our schemes between themselves and similar methods. As the tables show, both IBKRP and KRP-FS demonstrate good performances against their linear counterparts. We also find that linear rank pooling (RP), as well as BKRP are often out-performed by average pooling (AP) -- which is unsurprising given that the CNN features are non-linear (for RP) and the pre-image computed (as in BKRP) might not be a useful representation without the reconstructive term as in IBKRP or KRP-FS. We also find that KRP-FS is about 3-5\% better than its linear variant GRP on most of the datasets. 
\begin{table}[htbp]
\centering
\begin{tabular}{c|c|c|c|c}
RP & GRP & BKRP & IBKRP & KRP-FS\\
\hline
1.1 & 3.8 & 6.7 & 8.8 & 9.5
\end{tabular}
\caption{Avg. run time (time taken / frame) -- in milli-seconds -- on the HMDB dataset. CNN forward pass time is not included.}
\label{tab:runtime}
\end{table}

\begin{table}[htbp]
	\centering
	\begin{tabular}{c|c}
        \hline
		Kernel Sampling Factor &  Accuracy    \\		
		\hline
        1/32 & 60.56 \\
       	1/8  & 61.43 \\
        1/2  & 61.7 \\ 
	\end{tabular}
	\caption{Influence of Nystr\"om approximation to the KRP-FS kernel, using 3 subspaces on HMDB split-1.}         
    \label{tab:hmdb_nystrom}
\end{table}
%
%


\subsection{Comparisons to the State of the Art}
In Tables~\ref{tab:hmdb_soa},~\ref{tab:mpii_soa}, and~\ref{tab:jhmdb_soa},  we showcase comparisons to state-of-the-art approaches. Notably, on the challenging HMDB dataset, we find that our method KRP-FS achieves 69.8\% on 3-split accuracy, which is better to GRP by about 4\%. Further, by combining with Fisher vectors-- IDT-FV -- (using dense trajectory features), which is a common practice, we outperform other recent state-of-the-art methods. We note that recently Carreira and Zissermann~\cite{carreira2017quo} reports about 80.9\% accuracy on HMDB-51 by training deep models on the larger Kinectics dataset~\cite{kay2017kinetics}. However, as seen from Table~\ref{tab:hmdb_soa}, our method performs better than theirs (by about 3\%) when not using extra data. We outperform other methods on MPII and JHMDB datasets as well -- specifically, KRP-FS when combined with IDT-FV, outperforms GRP+IDT-FV by about 1--2\% showing that learning representations in the kernel space is indeed useful.

\subsection{Comparisons to Hand-crafted Features}
In Table~\ref{tab:utkinect-results}, we evaluate our method on the bag-of-features dense trajectories on MPII and non-linear features for encoding human 3D skeletons on the UTKinect actions. As is clear from the tables, all our pooling schemes significantly improve the performance of linear rank pooling and GRP schemes. As expected, IBKRP is better than BKRP by nearly 8\% on UT Kinect actions. We also find that KRP-FS performs the best most often, with about 7\% better accuracy on the MPII cooking activities dataset against GRP and UT Kinect actions. These experiments demonstrate the representation effectiveness of our method with regard to the diversity of the data features.

\begin{table}[htbp]
	\centering
	\begin{tabular}{c|c|c|c}
        \hline
		JHMDB Dataset &  FLOW  & RGB & FLOW+RGB \\				
		\hline
		Avg. ~\cite{simonyan2014two}         & 63.8 &  47.8 & 71.2  \\
        RP~\cite{Fernando2016} & 41.1 &  47.3 & 56.0  \\
        GRP~\cite{cherian_grp}     & 64.2 &  42.5 & 70.8 \\
        \hline
        BKRP   (ours)               & 65.8 &  49.3 & 73.4 \\
        IBKRP  (ours)               & 68.2 &  49.0 & \textbf{76.2}\\
        KRP-FS (ours)              & 67.5 &  46.2 & {74.6}\\
	\end{tabular}
	\caption{Classification accuracy on the JHMDB dataset split-1.}           
    \label{tab:jhmdb-split1-results}
\end{table}

\begin{table}[htbp]
	\centering
	\begin{tabular}{c|c|c|c}
        \hline
		HMDB Dataset &  FLOW  & RGB & FLOW+RGB \\				
		\hline
		Avg. Pool ~\cite{simonyan2014two}         & 57.2 & 45.2 & 65.6  \\
        RP~\cite{Fernando2016} & 56.7 &  38.3 & 63.1  \\
        GRP~\cite{cherian_grp}     & 65.3&  47.8 & 68.3 \\
        \hline
        BKRP   (ours)               & 54.9 &  45.9 & 69.5 \\
        IBKRP  (ours)               & 58.2 &  46.8 & 69.6\\
        KRP-FS (ours)              & 66.1 &  54.1 & \textbf{71.9}\\
	\end{tabular}
	\caption{Classification accuracy on the HMDB dataset split-1. }           
    \label{tab:hmdb-split1-results}
\end{table}
\begin{table}[htbp]
	\centering
	\begin{tabular}{c|c|c|c}
		MPII Dataset &  FLOW  & RGB & FLOW+RGB \\				
		\hline
		Avg. ~\cite{simonyan2014two}         & 48.1 & 41.7 & 51.1  \\
        RP~\cite{Fernando2016} & 49.0 &  40.0 & 50.6  \\
        GRP~\cite{cherian_grp}     & 52.1 &  50.3 & 53.8 \\
        \hline
        BKRP   (ours)               & 40.5 &  35.5 & 42.9 \\
        IBKRP  (ours)               & 52.1 &  43.2 & 55.9\\
        KRP-FS (ours)              & 48.2 &  44.7 & \textbf{57.2}\\
	\end{tabular}
	\caption{Classification accuracy (mAP\%) on the MPII dataset split1.}           
    \label{tab:mpii-split1-results}
\end{table}

\begin{table}[htbp]
	\centering
    \begin{tabular}{c|c}
		Algorithm &  Acc.(\%) \\				
		\hline
		Avg. Pool & 42.1\\
        RP~\cite{fernando2015modeling} & 45.3\\
        GRP~\cite{cherian_grp} & 46.1\\
        \hline
        BKRP   &  46.5\\
        IBKRP  &  49.5\\
        KRP FS  & \textbf{53.0}\\
        \hline
	\end{tabular}
    \quad
	\begin{tabular}{c|c}
	    Algorithm &  Acc.(\%) \\
        \hline
        SE(3) ~\cite{vemulapalli2014human} & 97.1\\
        Tensors~\cite{koniusz2016tensor} & 98.2\\		
        RP~\cite{fernando2015modeling} & 75.5\\   
        \hline
        BKRP  &  84.8\\
        IBKRP &  92.1\\    
        KRP FS & \textbf{99.0}\\
        \hline
	\end{tabular}
	\caption{Performances of our schemes on: dense trajectories from the MPII dataset (left) and UT-Kinect actions (right). For KRP-FS on UTKinect actions, we use 15 subspaces.}
      \label{tab:utkinect-results}
\end{table}   
\begin{table}[h]
	\centering
	\begin{tabular}{c|c}
		Algorithm &  Avg. Acc. (\%) \\
		\hline
        ST Multiplier Network\cite{feichtenhofer2017spatiotemporal}        & 68.9\%     \\
		ST Multiplier Network + IDT\cite{feichtenhofer2017spatiotemporal}        & 72.2\%       \\
        Two-stream I3D\cite{carreira2017quo}               & 66.4\% \\
        Temporal Segment Networks~\cite{Wang2016} & 69.4 \\
        Hier. Rank Pooling + IDT-FV~\cite{fernando2016discriminative} & 66.9\\        
        GRP  & 65.4 \\
		GRP + IDT-FV & 67.0\\
		\hline		
        BRKP &  64.1 \\
        IBKRP & 66.3 \\
        IBKRP + IDT-FV & 67.6 \\
        KRP-FS & 69.8\\
        KRP-FS + IDT-FV & \textbf{72.7}\\	
	\end{tabular}    
	\caption{HMDB Dataset (3 splits)}	
    \label{tab:hmdb_soa}    
\end{table}
\begin{table}[h]
	\centering    
	\begin{tabular}{c|c}
		Algorithm &  mAP(\%) \\
		\hline
		Interaction Part Mining~\cite{zhou2015interaction} & 72.4 \\
		Video Darwin~\cite{fernando2015modeling}    & 72.0 \\
		Hier. Mid-Level Actions~\cite{su2016hierarchical} & 66.8\\
        PCNN + IDT-FV~\cite{cheron2015p} & 71.4 \\
        GRP~\cite{cherian_grp} & 68.4 \\
        GRP + IDT-FV~\cite{cherian_grp} & {75.5}\\
		\hline
		BRKP &  66.3 \\
        IBKRP & 68.7 \\
        IBKRP + IDT-FV & 71.8\\
		KRP-FS  & 70.0\\
        KRP-FS + IDT-FV & \textbf{76.1}\\		
	\end{tabular}    
	\caption{MPII Cooking Activities (7 splits)}	
    \label{tab:mpii_soa}
\end{table}
\begin{table}[h]
	\centering
	\begin{tabular}{c|c}
		Algorithm &  Avg. Acc. (\%) \\
		\hline
		Stacked Fisher Vectors~\cite{peng2014action} & 69.03\\
        Higher-order Pooling~\cite{hok} & 73.3\\
        P-CNN + IDT-FV~\cite{cheron2015p} & 72.2 \\
        GRP~\cite{cherian_grp}  & 70.6 \\
		GRP + IDT-FV~\cite{cherian_grp} & {73.7}\\
		\hline       
		BRKP &  71.5 \\
        IBKRP & 73.3 \\
        IBKRP + IDT-FV & 73.5\\
		KRP-FS & 73.8 \\
        KRP-FS + IDT-FV & \textbf{74.2}\\				
	\end{tabular}    
	\caption{JHMDB Dataset (3 splits)}	
    \label{tab:jhmdb_soa}
\end{table}


\section{Conclusions}
\label{sec:conclude}
In this paper, we looked at the problem of compactly representing temporal data for the problem of action recognition in video sequences. To this end, we proposed kernelized subspace representations obtained via solving a kernelized PCA objective. The effectiveness of our schemes were substantiated exhaustively via experiments on several benchmark datasets and diverse data types. Given the generality of our approach, we believe it will be useful in several domains that use sequential data.

{\small
\bibliographystyle{ieee}
\bibliography{krp_bib}
}

\end{document}